\documentclass[10pt,twocolumn,letterpaper]{article}

\usepackage[pagenumbers]{cvpr} 
\usepackage{graphicx,amsmath,amsfonts,amssymb,caption,subcaption,multirow,overpic,textpos,makecell,url,booktabs,nicefrac,microtype,xspace,array, soul}

\usepackage[table]{xcolor}
\usepackage[british, american]{babel}
\usepackage[pagebackref,breaklinks,colorlinks]{hyperref}

\makeatletter
\@namedef{ver@everyshi.sty}{}
\makeatother
\usepackage{tikz}

\usepackage{pifont}
\usepackage{bbding}
\usepackage{wasysym}
\usepackage{utfsym}
\usepackage{fontawesome}

\usepackage{color}

\usepackage[capitalize]{cleveref}
\crefname{section}{Sec.}{Secs.}
\Crefname{section}{Section}{Sections}
\Crefname{table}{Table}{Tables}
\crefname{table}{Tab.}{Tabs.}

\newlength\savewidth\newcommand\shline{\noalign{\global\savewidth\arrayrulewidth
  \global\arrayrulewidth 1pt}\hline\noalign{\global\arrayrulewidth\savewidth}}
\newcommand{\tablestyle}[2]{\setlength{\tabcolsep}{#1}\renewcommand{\arraystretch}{#2}\centering\footnotesize}
\renewcommand{\paragraph}[1]{\vspace{1.25mm}\noindent\textbf{#1}}

\newcolumntype{x}[1]{>{\centering\arraybackslash}p{#1pt}}
\newcolumntype{y}[1]{>{\raggedright\arraybackslash}p{#1pt}}
\newcolumntype{z}[1]{>{\raggedleft\arraybackslash}p{#1pt}}

\newcommand{\app}{\raise.17ex\hbox{$\scriptstyle\sim$}}

\definecolor{deemph}{gray}{0.6}
\newcommand{\gc}[1]{\textcolor{deemph}{#1}}

\definecolor{baselinecolor}{gray}{.9}
\newcommand{\baseline}[1]{\cellcolor{baselinecolor}{#1}}

\def\cvprPaperID{} \def\confName{CVPR}
\def\confYear{2023}

\begin{document}

% \vspace{-.3in}
\title{Towards 3D Object Detection with 2D Supervision}

\vspace{-.1in}
\author{Jinrong Yang$^1$~~~~~ Tiancai Wang$^2$~~~~~ Zheng Ge$^2$~~~~~ Weixin Mao$^2$\\
Xiaoping Li$^1$~~~~~ Xiangyu Zhang$^2$\\
$^1$Huazhong University of Science and Technology~~~~ $^2$MEGVII Technology}
% \hspace{-.15in}
% \tt\small \{yangjinrong,}@hust.edu.cn;

\maketitle

\begin{abstract}
The great progress of 3D object detectors relies on large-scale data and 3D annotations.
The annotation cost for 3D bounding boxes is extremely expensive while the 2D ones are easier and cheaper to collect.
In this paper, we introduce a hybrid training framework, enabling us to learn a visual 3D object detector with massive 2D (pseudo) labels, even without 3D annotations.
To break through the information bottleneck of 2D clues, 
we explore a new perspective: \textbf{Temporal 2D Supervision}. We propose a temporal 2D transformation to bridge the 3D predictions with temporal 2D labels. Two steps, including homography wraping and 2D box deduction, are taken to transform the 3D predictions into 2D ones for supervision. Experiments conducted on the nuScenes dataset show strong results (nearly 90\% of its fully-supervised performance) with only 25\% 3D annotations. We hope our findings can provide new insights for using a large number of 2D annotations for 3D perception.

\end{abstract}

\section{Introduction\label{sec:intro}}

As a fundamental topic in 3D perception scenes, 3D object detection aims to detect a set of objects in 3D space with 3D-oriented boxes and corresponding categories. Due to the application of deep learning, both LiDAR-based and camera-based frameworks have achieved remarkable progress and superior performance. However, launching such cutting edges is ignited by \emph{large-scale} data and \emph{precise} 3D labels. The process of creating an application-level dataset undergoes extremely expensive resources.

To tackle this issue, the weakly supervised learning framework is an economic way to learn a 3D detector. Recent works focus on designing more economical labels to train models instead of expensive 3D labels. Most efforts focus on point-cloud scenes which leverage 2D bounding boxes~\cite{FGR} or reference BEV points~\cite{weak1} to supervise a model for generating coarse proposals. Then a point segmentation model is further employed to deduce the foreground points cloud for the subsequent refinement of 3D bounding boxes. These methods need complex multi-stage pipelines. Besides, WeakM3D\cite{WeakM3D} explores training the camera-only 3D detector with 2D labels and point cloud proposals. These efforts dramatically reduce the annotation cost, making it easy to scale the data.

However, the above approaches still expose several downsides: 1) both sensor data \ie, LiDAR point cloud, and image, are required for learning a detector of a single-modal; 
2) the training pipelines are sophisticated since they usually train with multiple stages and require several independent algorithms (\eg, RANSAC~\cite{fischler1981random} to remove ground parts, and extra point segmentation model, \etc). To this end, one question may arise: 
\emph{could we leverage massive 2D labels (\ie, 2D boxes and category labels) to train a 3D camera-based detector in an end-to-end way without involving point clouds?}

\begin{figure}[t]
\centering
\hspace{-.04in}
\includegraphics[width=0.48\textwidth]{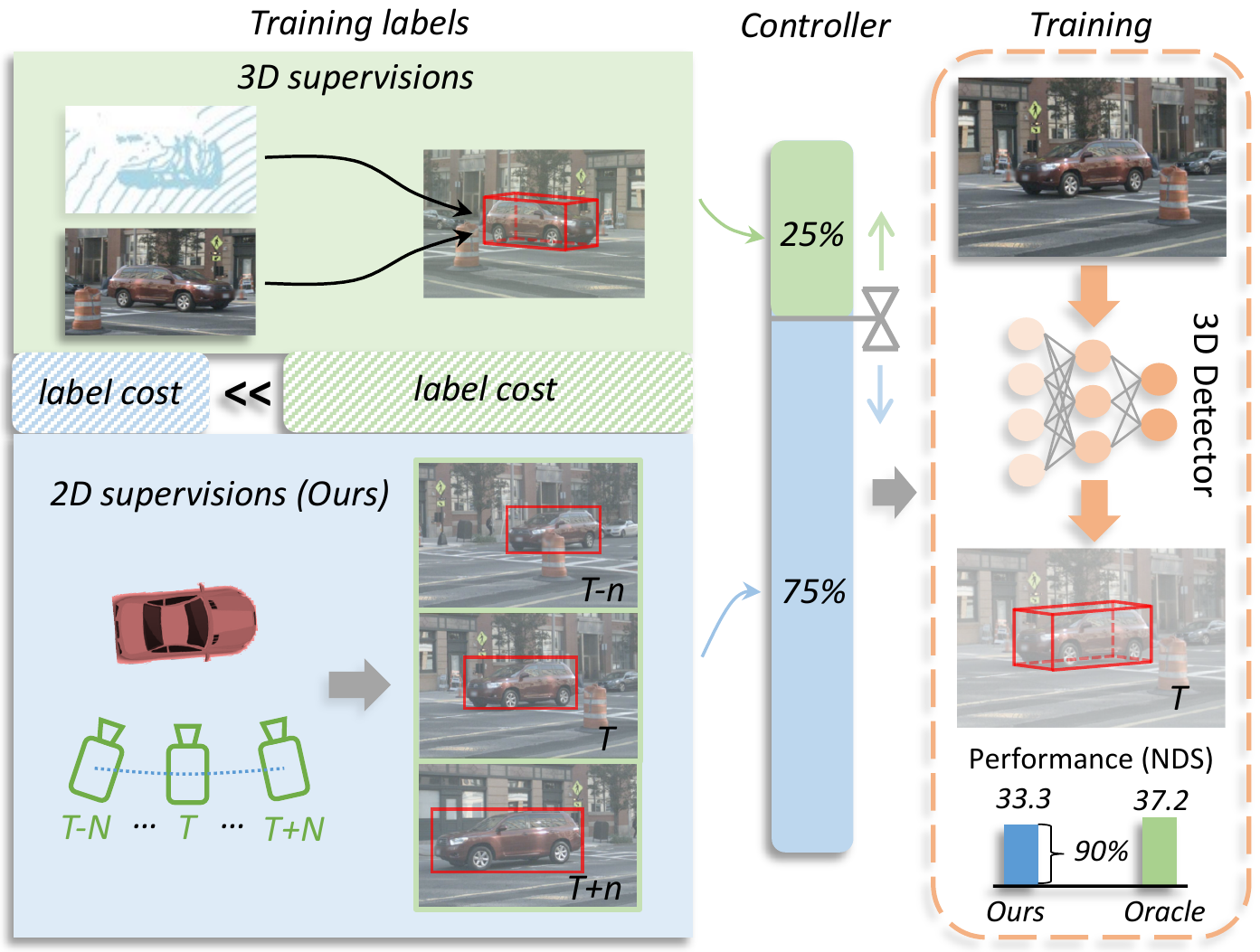}\\
\caption{\textbf{Illustration on the cost 3D and our cheaper 2D supervisions.} 3D labels need to adjust the 3D box referring to both LiDAR point cloud and image information while our method only requires temporal 2D boxes without depending on point cloud. Our method is also flexible for controlling the ratio between two supervisions. Only using 25\% strong supervision, our method achieves 90\% performance of full 3D supervision.}
\label{fig:motivation}
\vspace{-.15in}
\end{figure}

Training a 3D detector with 2D annotations is \emph{challenging} yet \emph{promising}. Its challenge is reflected by the information bottleneck of 2D clues (\eg, depth, size and rotation).
Whilst it is also promising since some precedents provide valuable insights to deduce 2D clues for learning 3D representation.
A classical application is Multi-View Stereo (MVS)~\cite{mvsnet,scharstein2002taxonomy}, which poses multiple spatial or temporal views to reconstruct 3D scenes with only 2D images.
Another advanced application is differentiable rendering~\cite{renderer,kato2018neural,liu2019soft}. 
It also leverages 2D image supervision of different views to render a single 3D object. 
Going into one step, \cite{sul} extents the differentiable rendering to learn the 3D shapes and layouts of multiple objects with only 2D dense masks. 
Inspired by those advances in MVS and differentiable rendering, we explore 3D object detection by digesting the 2D supervision at scale.

In this paper, we aim to develop a hybrid training framework (see Fig.~\ref{fig:motivation}) that learns a 3D visual detector with massive 2D labels and a few 3D annotations, even \emph{without 3D labels}. To digest the 2D supervision in the 3D detector, we explore a new perspective to learn 3D representation: \emph{temporal 2D supervision}. Our hybrid framework can be built upon any 3D visual detectors (\eg, FCOS3D~\cite{fcos3d}). We first use a homography matrix to wrap the predicted 3D boxes to the adjacent frames. Afterwards, we introduce a non-parametric projection to deduce the warped 3D boxes to 2D ones, which can be directly supervised by temporal 2D (pseudo) labels. With temporal 2D supervision, 3D information (\eg, depth, size, and rotation) can be learned implicitly. 
With the novel pipeline, we further observe some corner cases introduced by the moving objects and propose some regularization strategies:
1) symmetrical temporal supervision and 2) appropriate temporal intervals. 

We conduct comprehensive experiments on nuScenes dataset~\cite{nuScenes} to show the effectiveness of our method. Experimental result shows that strong performance (nearly 90\% of its fully-supervised performance) are achieved with only 25\% 3D annotations. It's worth noting that our method with 100\% 2D supervision can produce a promising result (60\% of its 3D oracle performance). 
With our training framework, it is flexible to use any ratio between 2D and 3D labels, enabling control of the annotation cost in practical applications.
We hope our methodology and studies can provide new insights for using large-scale 2D labels for 3D detectors.

\section{Related Work\label{sec:related}}

\paragraph{3D object detection} focus on detecting objects with 3D-oriented boxes in 3D space. With the great progress in deep learning, LiDAR-based and camera-based 3D detectors are continuously evolving. One stream of LiDAR-based works~\cite{pointrcnn,3dssd,aissd,dbqssd} directly processes vanilla point cloud in 3D space and leverages PointNet++-like network to encode features for prediction. Several works~\cite{second,pointpillars,voxelnet,centerpoint} focus on splitting point cloud into regular voxels followed by pooling to 2D representation, after which 2D convolution networks are naturally utilized to process feature maps. In addition, recent efforts aim to conduct 3D object detection tasks based on image modal, which is a more cheap scheme. They employ a 2D backbone to extract feature maps and attach a 3D target-specific head to predict objects in 3D space. CenterNet~\cite{centernet} introduces a heatmap-based head to perceive objects and eliminate the NMS procedure. Inspired by dense prediction framework in 2D tasks~\cite{fcos,retinanet}, FCOS3D~\cite{fcos3d} predicts objects in each location of the multi-level feature map. The great breakthroughs made by the above 3D detector benefit from creating large-scale datasets with precise 3D annotations. The process of collecting such a dataset undergoes time-consuming and expensive resources so we explore leveraging cheaper 2D labels for training in this paper. 

\paragraph{Weakly supervised 3D object detection} is a promising way to economically train models. It aims to leverage the partial representation of 3D labels \eg, 2D boxes, and BEV points, to learn a 3D detector. Recently, Meng \emph{et al.}~\cite{weak1} proposes to annotate a small part of BEV points and precise 3D labels in LiDAR point cloud data for training. It designs a two-stage architecture for predicting results so a two-stage training process is also conducted. To eliminate the requirement of 3D labels, FGR~\cite{weak3} proposes to leverage only 2D boxes in the image to project coarse 3D point cloud space. Then a point segmentation model is used to identify foreground points for the subsequent 3D box prediction. The weakly supervision technologies are well explored in LiDAR-based frameworks but rarely involve camera-based ones until the coming of WeakM3D~\cite{WeakM3D}. It similarly employs 2D boxes in image space to crop point cloud proposals which is leverage to build strong constraints for predictions of cameras. While these works provide several advanced methods for cheap training detectors, they still undergo two issues: 1) both modals need not be absent for carrying out supervision; 2) the training pipeline requires multiple stages and complex algorithms.
In this paper, we not only utilize temporal 2D labels for training a camera-based detector without LiDAR point cloud but train our model in an end-to-end manner without extra complex algorithms.

\begin{figure*}[t]
\centering
\vspace{-.1in}
\includegraphics[width=1.0\textwidth]{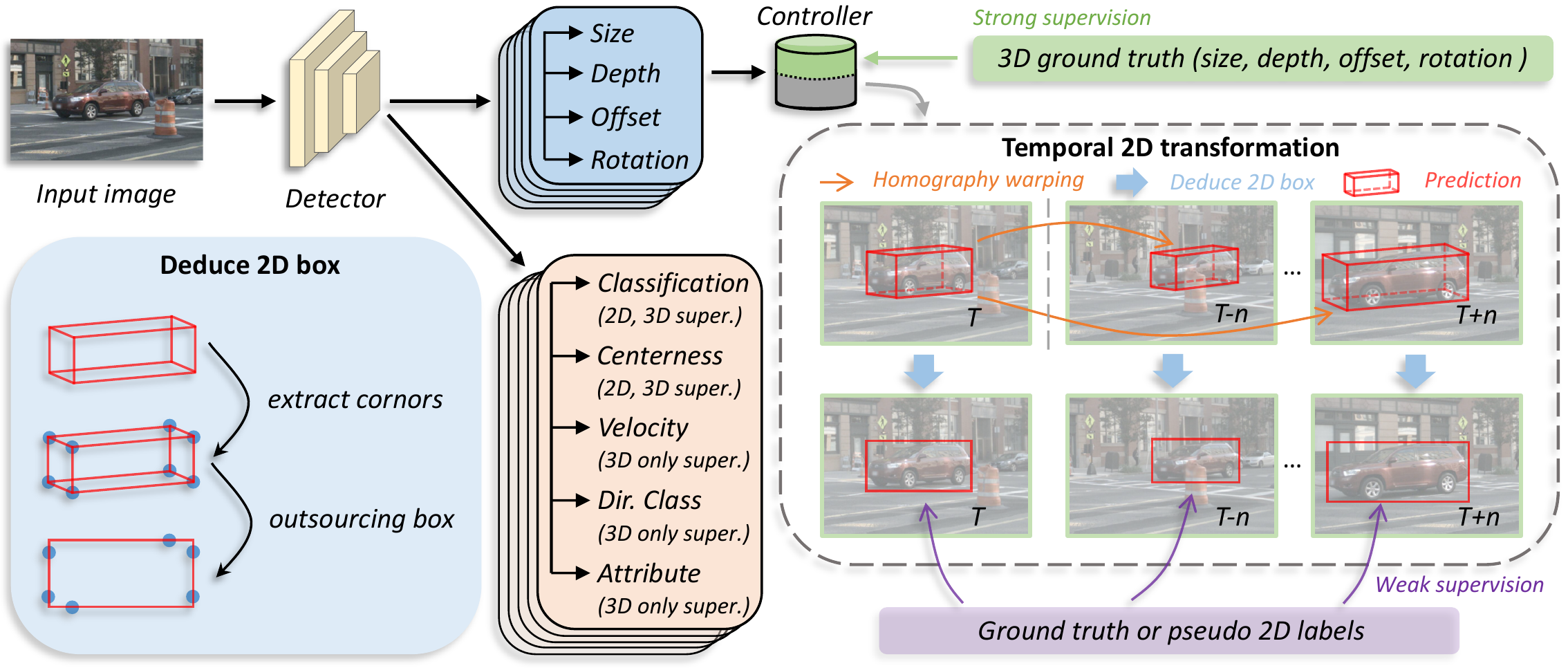}\\
\vspace{-.1em}
\caption{\textbf{Overall framework.} The current frame image is input to the monocular detector for predicting 3D representation and classification. During training, a controller is utilized to adjust the ratio between strong and weak supervision. The strong supervision part directly adopts precise 3D ground truth for building loss, while the weak part needs to carry out our proposed transformation for generating predicted temporal 2D representations. The transformation is parameter-free, including the homography warping and the 2D box deduction processes. Therefore, ground truth (pseudo) labels can implicitly supervise 3D representations. The supervision of the classification and center-ness branches is conducted in both strong and weak manners. The velocity, direction, and attribute labels are only usable under strong supervision.}
\label{fig:approach}
\vspace{-.15in}
\end{figure*}

\paragraph{Learning 3D tasks with 2D supervision} is also a promising topic in many tasks like Multi-View Stereo (MVS), differentiable rendering, and 3D shape and layout perception \etc. By structuring projection between 3D and 2D representations, multi-view or temporal 2D cues may be used to supervise the training process for the 3D perception tasks. MVS technology~\cite{mvsnet,scharstein2002taxonomy} structures a cost volume among multi-view 2D images and further measures similarity for 3D object structure reconstruction. Differentiable rendering~\cite{renderer,kato2018neural,liu2019soft} focuses on leveraging 2D representations to flow information to 3D visuals for rendering object details. Recent work~\cite{sul} delves into the 3D shape and layout prediction framework by training its model with multi-view 2D images. It successfully learns dense depth information and layout in 3D space.
Motivated by these precedents, we explore training the monocular 3D detector by temporal 2D supervision. To the best of our knowledge, it is the first effort to train a camera-based detector with only 2D supervision, which is non-trivial to build projection between sparse 3D and 2D representations.

\section{Preliminary}

Our framework can be built on any vision detector.
In this paper, we mainly explore the hybrid framework based on the monocular 3D detector FCOS3D~\cite{fcos3d}.

Before stepping into our method, we first review the pipeline of FCOS3D~\cite{fcos3d}.
It keeps the fully convolutional network for 2D detection tasks and is applied for 3D object detection by extending the 3D target head. 
As shown in Fig.~\ref{fig:approach}, an image is first processed by a backbone~\cite{resnet} followed by a feature pyramid network (FPN)~\cite{fpn}, which extracts multi-scale feature maps with five different resolutions.
Each level of FPN feature map predicts the parameters of 3D box and the corresponding class for each location.
It tiles dense anchor points on feature maps with multiple scales and designs a center-based strategy to select the positive and negative samples.
Then, two parallel branches are attached for classification and regression, respectively.
The classification branch predicts the category or background for each location.
As for the regression branch, it predicts 3D representation of all positive locations, involving 3D size $(w, l, h)$, offsets $(\Delta x, \Delta y)$ to the projected 3D center, normalized depth $d$, rotation $\theta$, direction class $C_\theta$, and 3D aware center-ness $c$.

\section{Method \label{sec:approach}}

Our method is a hybrid framework for training a camera-only 3D detector with massive cheap 2D labels and a few 3D annotations in an end-to-end manner. 
For 2D supervision, we adopt temporal 2D bounding boxes for training and attempt to endow the model with the ability to predict 7-DoF 3D boxes in 3D space. While for 3D supervision, we adopt the common way in 3D object detectors. 

\subsection{Motivation}\label{sec:method:motivation}

Learning 3D object detectors with 2D supervision is economical yet challenging due to the lack of valid depth, 3D size, or rotation information.
Recent works~\cite{WeakM3D,FGR} aim to remedy these incomplete information with the aid of point clouds.
By fetching the coarse foreground point clouds projected from 2D supervision, the information from the point cloud can be adopted to supervise 3D predictions.
Though these approaches are effective, they still require LiDAR data for training.

In this paper, we explore training vision 3D detectors with pure 2D labels.
Recent precedents~\cite{mvsnet,scharstein2002taxonomy,renderer,kato2018neural,liu2019soft,sul} leverage 2D clues to learn a 3D perception task, \eg, 3D shape and layout estimation.
By structuring the mapping or constraint conditions between 3D and \textbf{multi-view} 2D representations, the 3D predictions can be projected to 2D space, supervised by 2D labels.
Motivated by these priors, we aim to perform 3D object detection by the supervision of multi-view 2D boxes.
However, it is difficult to construct multi-view 2D boxes for supervision in a single frame.
We find using the \textbf{temporal} 2D boxes to build the multi-view boxes is a key factor to achieve our goal.
Fortunately, the dataset and its 3D labels are collected and annotated with video streaming.
Therefore, we can simply deduce the 2D boxes from 3D labels for supervision.
After meeting the supervision requirement, we also need to design a \textbf{bridge} between the 3D predictions and the temporal 2D supervisions.
To this end, we introduce a non-parameter transformation, including homography warping and 2D box deduction, on the 3D predictions.
Such transformation will be detailed in the next section.

\subsection{Framework Overview}\label{sec:method:framework}

In this section, we demonstrate the overall learning pipeline of our method supervised by massive temporal 2D labels and a few 3D labels.
As shown in Fig.~\ref{fig:approach}, the 3D detector normally predicts 3D boxes.
During training, we design a controller to adjust the proportion between 3D labels and temporal 2D labels for supervision.
The 3D supervision part inherits the learning approach of FCOS3D.
For 2D supervision, we leverage the non-parameter transformation to transform the predicted 3D boxes into \emph{temporal} 3D space and deduce them into 2D ones (see Sec.~\ref{sec:method:transform}).
By directly employing ground truth (pseudo) 2D labels to supervise these transformed 2D boxes, the representations of 3D predictions can be learned \emph{implicitly}.
Our method is trained in an end-to-end manner and does not need any other clues \eg, LiDAR point cloud, to impose complex constraints.

\begin{figure}[t]
\centering
% \hspace{-.1in}
\includegraphics[width=0.43\textwidth]{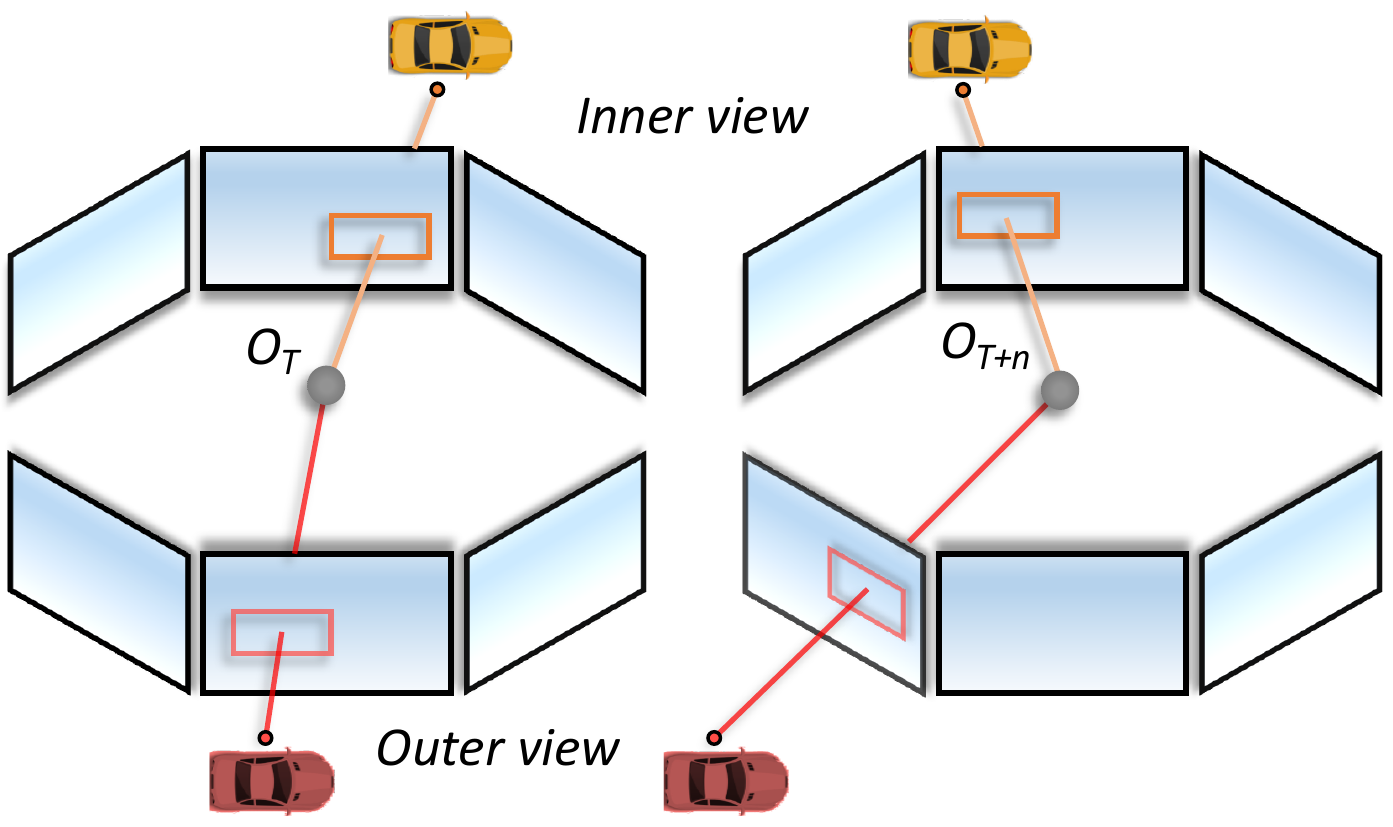}\\
\caption{\textbf{Two types of 2D supervision.} (1) the temporal 2D supervision of the yellow car appear in the same views between training and supervised samples; (2) The red car appears in two different views so the pose between them needs to be used.}
\label{fig:span}
\vspace{-.1in}
\end{figure}

\subsection{Temporal 2D Transformation}\label{sec:method:transform}

In this section, we introduce the temporal 2D transformation to transform the predicted 3D boxes into temporal 2D ones.
It is a key condition for supervision with temporal 2D ground truths.
As shown in Fig.~\ref{fig:approach}, it mainly includes two steps: homography wrapping and deducing 2D boxes.

\paragraph{Homography wrapping.}
Given an input image $I_v^t$ at frame $t$ from one camera view $v \in {V}$\footnote{$V$ is \{\texttt{CAM\_FRONT}, \texttt{CAM\_FRONT\_RIGHT}, \texttt{CAM\_FRONT\_LEFT}, \texttt{CAM\_BACK}, \texttt{CAM\_BACK\_LEFT}, \texttt{CAM\_BACK\_RIGHT}\} in nuScenes dataset.}, the detector predicts a set of 3D boxes $\{B_i^t\}_{i=1}^N$ in camera coordinate.
Each 3D box is represented by 7-DOF parameters $(x_i, y_i, z_i, w_i, h_i, l_i, \theta_i)$, which are center coordinates, size (width, height, and length), and rotation, respectively.
We use the homography matrix to warp the predicted box $B_i^t$ of the current frame $t$ to the camera coordinate space of adjacent frame $B_i^{t+\Delta t}$, where $\Delta t$ indicates the temporal interval between two frames.
The homography wrapping process can be formulated as:
\begin{equation}
    B_i^{t+\Delta t} = H_{t \rightarrow t+\Delta t} \cdot B_i^t,
\end{equation}
where $H_{t \rightarrow t+\Delta t}$ is the homography matrix of camera coordinate from $t$ to $t+ \Delta t$.
Such homography transformation may involve two typical cases: the inner view and outer view (see Fig.~\ref{fig:span}). It decides that we need to build different homography matrices:
\begin{equation}
\begin{cases}
 H_{t \rightarrow t+\Delta t} = C_t^{V(t)} \cdot P_t \cdot P_{t+\Delta t}^\top \cdot {C_{t+\Delta t}^{V(t+\Delta t)}}^\top
 \\ 
 % \\ 
 C_t^{V(t)} = C_{t+\Delta t}^{V(t+\Delta t)}, \qquad \qquad \text{ if Inner view}
 \\
 % \\
 C_t^{V(t)} \neq C_{t+\Delta t}^{V(t+\Delta t)}, \qquad \qquad \text{ if Outer view}
\end{cases},
\end{equation}
where $C_t^{V(t)} \in \mathbb{R}^{4\times4}$ is the homogeneous matrix from the image view $V(t)$ of camera coordinate to the ego coordinate at time $t$.
$P_t \in \mathbb{R}^{4\times4}$ indicates the pose from ego to global coordinate.
Both $C$ and $P$ include translation and rotation transformations.
Therefore, transforming the predicted 3D boxes of current frame $t$ to adjacent frame $t+\Delta t$ is a homography warping procedure.

\paragraph{Deduce 2D box.}
After warping the 3D predictions into the camera coordinates of the adjacent frame, here we introduce how to project the 3D representations to 2D boxes.
The blue region in the bottom left corner of Fig.~\ref{fig:approach} shows the projection in detail.
Inspired by PGD~\cite{pgd}, we project the 3D representations to 2D space by extracting the outsourcing box in the same image perspective.
Given a warped 3D box $B_i^{t+\Delta t} \in \mathbb{R}^{7}$ in a view $V(t+\Delta t)$ at time $t+\Delta t$, we first extract the eight corner points $O_i^{V(t+\Delta t)} \in \mathbb{R}^{8 \times 3}$ in camera coordinate.
Then, the corresponding camera’s intrinsic parameter $K(t+\Delta t) \in \mathbb{R}^{4\times 4}$ is used to map corners to the image plane:
\begin{equation}
    O_i^{pix} = K(t+\Delta t) \cdot O_i^{V(t+\Delta t)},
\end{equation}
where $O_i^{pix} \in \mathbb{R}^{8\times2}$ is the pixel coordinate of projected corners.
Afterwards, we calculate the leftmost, rightmost, topmost, and downmost corners for constructing 2D box $D_i \in \{x_{tl}, y_{tl}, x_{br}, y_{br}\}$.

The overall temporal 2D transformation is non-parametric. During training, the predicted 3D boxes can be transformed into 2D boxes of arbitrary adjacent frames, \ie the $\Delta t$ can be arbitrarily positive or negative values.

\begin{figure}[t]
\centering
\hspace{-.05in}
\includegraphics[width=0.48\textwidth]{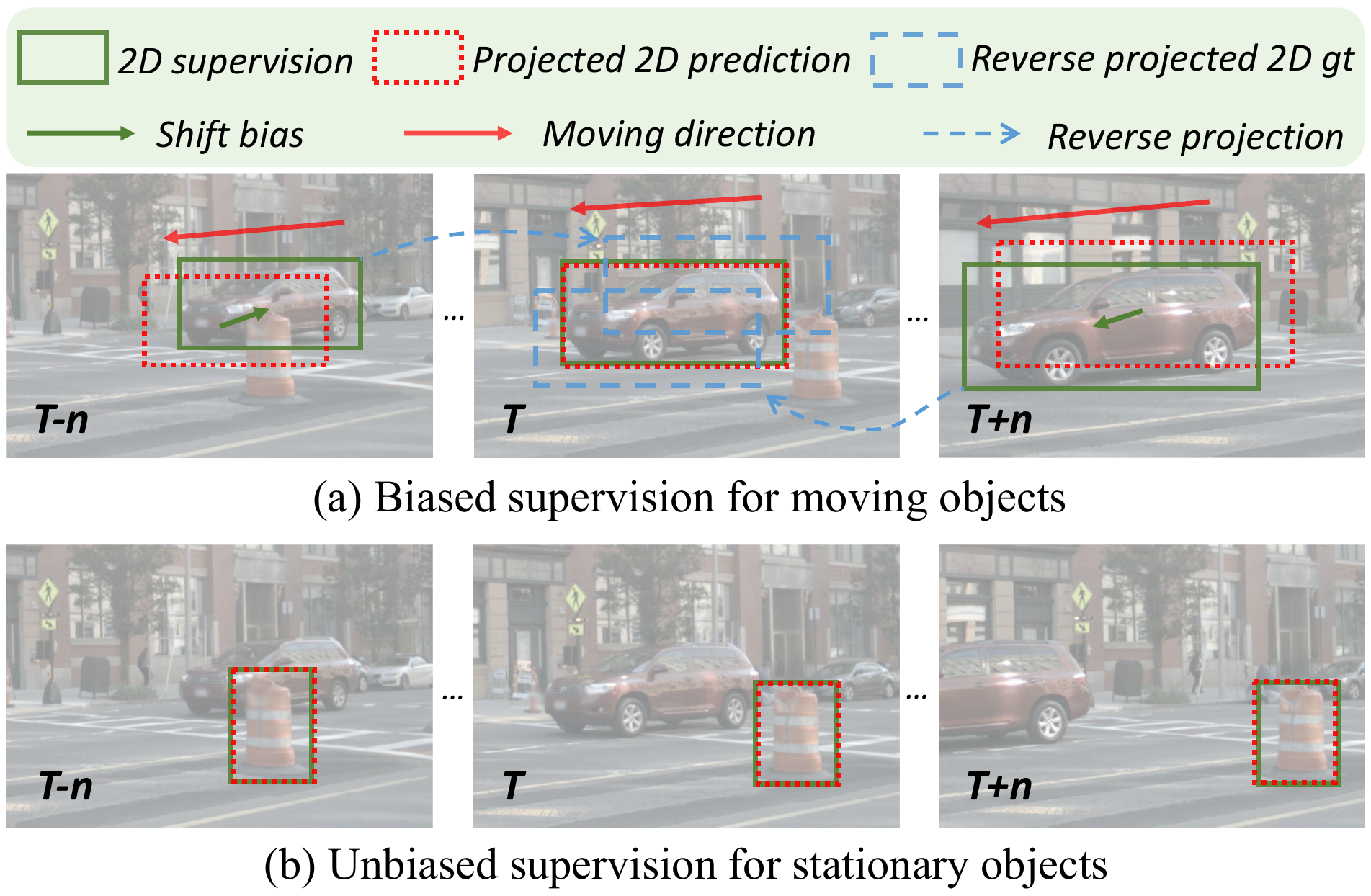}\\
% \vspace{-.1em}
\caption{\textbf{Motion ambiguity.} (a) biased supervision appears in moving objects since it suffers from location shift; (b) the supervision for stationary objects is unbiased since it is a homography warping among temporal frames.}
\label{fig:supervision}
\vspace{-.1in}
\end{figure}

\subsection{Motion ambiguity}

In this section, we reveal that temporal 2D supervision is prone to fall into suboptimal training. We traced the problem to a culprit: \emph{motion ambiguity}. Afterwards, we introduce two strategies to regularize it.

\paragraph{How does motion ambiguity affect training?} We start out analyzing this issue by probing the instances' motion distribution of nuScenes dataset~\cite{nuScenes}. We find there are 26\% objects in motion state which is actually the culprit.
As shown in Fig.~\ref{fig:span} (a), it indicates that the GT 2D boxes in temporally adjacent frames can match the objects. However, by reversing them into the current frame $T$, it triggers \emph{location and size misalignment} with the same object (GT box). The issue is attributed to the motion of objects, making ambiguity to estimate the 3D information~\cite{manydepth,megnet,bevstereo} with the homography warping transformation. Therefore, it will cause bias supervision with temporal 2D labels.

\paragraph{Two regularization strategies.}
To alleviate this issue, we find that the object's motion normally equips with spatiotemporal consistency~\cite{adascale,stream}, \ie, the moving tendency of location shift and direction are consistent during a continuous period.
Based on this insight, we use \emph{symmetrical temporal 2D supervision} (\eg, $\Delta t = \{-\tau, 0, \tau\}$) for constructing loss.
It can effectively offset the bias among the opposite optimizations.
As for the 2D labels in the current frame and all stationary objects, such bias issue is not observed (see Fig.~\ref{fig:span} (b)).
In addition, we empirically find that \emph{reasonably large perspective} (\eg, $\Delta t$ is set as $\{-3, 0, 3\}$) achieves the best performance in our practice.
The possible reason is that short temporal intervals may not introduce enough parallax for depth estimation and even bring about some noise for supervision.
The effectiveness of both strategies will be further verified in Tab.~\ref{tab:temporal_2d_supervision}.

\subsection{Hybrid Training}\label{sec:joint_training}
In this section, we introduce how to train the 3D detector in a hybrid manner with any proportion of 2D labels.
To achieve it, we need to formulate the label assignment for both 2D and 3D supervision, which is important~\cite{fcos,atss,ota} for dense prediction frameworks like FCOS3D~\cite{fcos3d}. 

\paragraph{Label assignment.}
For both 2D and 3D supervision, we use the center distance prior to assign positive and negative samples, similar to FCOS3D.
For the 3D part, FCOS3D adopts the projected 2D center of 3D ground truth in the image plane for reference. 

For pure 2D supervision, 2D ground-truth (GT) is short of depth information for projection. We employ the center of 2D labels to replace the projected 2D center, which has little impact on performance (see the ``w/o 3D LA'' in Tab~\ref{tab:remove_3d_supervision}).
In addition, the GT center-ness is also derived from the 2D box center in a 2D Gaussian distribution manner. For more label assignment details, please refer to \cite{fcos3d}. 

\paragraph{Supervision.} The overall loss function is formulated as:
\begin{equation}
    \mathcal{L}_{hybrid}=\lambda_{3D} \mathcal{L}_{3D} + \lambda_{2D} \mathcal{L}_{2D},
\end{equation}
where the $\lambda_{2D}$ and $\lambda_{3D}$ are the loss weights for the 2D and 3D supervisions, respectively. For the $\mathcal{L}_{3D}$, we follow the same loss function as FCOS3D:
\begin{equation}
\begin{aligned}
    \mathcal{L}_{3D}= & \frac{1}{N_{pos}}(\alpha_{cls}\mathcal{L}_{cls}+\alpha_{attr}\mathcal{L}_{attr}\\
    &+\alpha_{loc}\mathcal{L}_{loc}^{3D}+\alpha_{dir}\mathcal{L}_{dir}+\alpha_{ct}\mathcal{L}_{ct}^{3D}),
\end{aligned}
\end{equation}
where the focal loss and binary cross entropy (BCE) loss are used for the classification branch, and center-ness regression denoted as $\mathcal{L}_{cls}$, $\mathcal{L}_{ct}^{3D}$, respectively. Softmax classification loss is used for direction and attribution classifications denoted as $\mathcal{L}_{dir}$ and $\mathcal{L}_{attr}$. Smooth L1 loss is adopted to regress $(\Delta x, \Delta y, w, l, h, d, v_x, v_y)$.

For 2D supervision, we adopt the temporal 2D labels $D^{t+\Delta t}$ to supervise the transformed 2D predictions $\widehat{D}^{t+\Delta t}$:
\begin{equation}
    \mathcal{L}_{2D}=\frac{1}{N_{pos}}(\beta_{cls}\mathcal{L}_{cls}+\beta_{loc}\mathcal{L}_{loc}^{2D}+\beta_{ct}\mathcal{L}_{ct}^{2D}),
\end{equation}
% We use the GIoU~\cite{giou} to build loss.
% Therefore, the objective regression loss is:
where $N_{pos}$ is the number of positive samples.  $\mathcal{L}_{cls}$, $\mathcal{L}_{ct}^{2D}$ are the classification and center-ness losses, respectively. $\beta_{cls}$, $\beta_{loc}$ and $\beta_{ct}$ are the corresponding loss weights. $\mathcal{L}_{loc}^{2D}$ is the localization loss 
and we adopt the GIoU~\cite{giou} for supervision:
\begin{equation}
    \mathcal{L}_{loc}^{2D}=\sum_{i=0}^{N}\sum_{j \sim \Delta t}\mathrm{GIoU}(\widehat{D}_i^j, D_i^j),
\end{equation}
where $\widehat{D}_i^j$ is the transformed 2D prediction and ${D}_i^j$ is temporal 2D label. $\Delta t \in \{-\tau, ..., 0, ... ,\tau\}$ is a temporal series. N is the total number of positive predictions. $\tau$ can be instanced to arbitrarily temporal 2D labels.

\begin{table*}[t]
\centering
\hspace{-1.em}
\subfloat[
\textbf{Strong oracle}. ($\dag$: ground truth depth for projecting 2D supervision)
\label{tab:strong_oracle}
]{\begin{minipage}{0.50\linewidth}
\begin{center}
\small
\tablestyle{4pt}{1.2}
\begin{tabular}{l|ccccccc}
Seq. ID & mAP$\uparrow$ & mATE$\downarrow$ & mASE$\downarrow$ & mAOE$\downarrow$ & mAVE$\downarrow$ & mAAE$\downarrow$ & NDS$\uparrow$ \\ 
\shline
\gc{-} & \gc{0.255}  & \gc{0.827} & \gc{0.271} & \gc{0.608} & \gc{1.309} & \gc{0.173} & \gc{0.339}\\
\baseline{0} & \baseline{0.259} & \baseline{0.837} & \baseline{0.271} & \baseline{0.576} & \baseline{1.334} & \baseline{0.175} & \baseline{0.344}\\
0$\dag$ & \textbf{0.260} & 0.829 & \textbf{0.269} & \textbf{0.561} & \textbf{1.309} & 0.172 &\textbf{0.346}\\
-1,0,1 & 0.250 & 0.852 & 0.272 & 0.644 & 1.197 & \textbf{0.163} & 0.332\\
-2,0,2 & 0.251 & \textbf{0.817} & 0.275 & 0.621  & 1.165 & 0.187 & 0.336 \\
-3,0,3 & 0.255 & 0.825 & 0.271 & 0.641 & 1.143 & 0.181 & 0.336\\
\end{tabular}
\end{center}
\end{minipage}
}
\hspace{2.5em}
\subfloat[
\textbf{Removing 3D supervision.} (LA: 3D aware label assign)
\label{tab:remove_3d_supervision}
]{\begin{minipage}{0.40\linewidth}
\begin{center}
\small
\tablestyle{4pt}{1.2}
\begin{tabular}{cl|ccccc}
 & Remove Sup. & mAP$\uparrow$ & mATE$\downarrow$ & mASE$\downarrow$ & mAOE$\downarrow$\\ 
\shline
A & - & 0.259 & 0.837 & 0.271 & 0.576\\
B & A w/o 3D LA & 0.255 &0.836  & 0.273 & 0.645\\
C & B w/o Size & 0.252 & 0.842 & 0.418 & 0.642 \\
D & C w/o Rotation & 0.253 & 0.834 & 0.548 & 1.486 \\
E & D w/o Offset & 0.252 & 0.840 & 0.552 & 1.479  \\
\baseline{F} & \baseline{E w/o Depth} & \baseline{0.026} & \baseline{1.149} & \baseline{0.856} & \baseline{1.187} \\
\end{tabular}
\end{center}
\end{minipage}
}
\\[2mm]
\hspace{-.6em}
\subfloat[
\textbf{Temporal 2D supervision.}
\label{tab:temporal_2d_supervision}
]{\begin{minipage}{0.6\linewidth}
\begin{center}
\small
\tablestyle{4pt}{1.2}
\begin{tabular}{l|cccccccccc}
 & A & B & C & D & E & F & G & H & I & J\\
Seq. ID & 0 & [0,1] & [0,2] & [0,3] & [-1,0,1] & [-2,0,2] & \baseline{[-3,0,3]} & [-4,0,4] & [-2$\sim$2] & [-3$\sim$3] \\ 
\shline
mAP$\uparrow$ & 0.026 & 0.121  &  0.128 &  0.123 & 0.124 & 0.134  &\baseline{\textbf{0.151}} & 0.146 &0.130  & 0.135\\
mATE$\downarrow$ & 1.149 & 0.990 & 0.969  & 1.013  & 0.984  & 0.977 & \baseline{\textbf{0.956}} & 0.960 & 0.971  & 0.970 \\
mASE$\downarrow$ & 0.856 & 0.574 &  0.538 & 0.552  & 0.550  & 0.555  & \baseline{0.556} & 0.557 & 0.548  & \textbf{0.535}\\
\end{tabular}
\end{center}
\end{minipage}
}
\hspace{4.2em}
\subfloat[
\textbf{Removing 3D prior.}
\label{tab:remove_3d_prior}
]{\begin{minipage}{0.27\linewidth}
\begin{center}
\small
\tablestyle{4pt}{1.2}
\begin{tabular}{l|ccc}
Jit Scale & mAP$\uparrow$ & mATE$\downarrow$ & mASE$\downarrow$\\ 
\shline
-  & 0.151 & 0.956 & 0.556 \\
0.05 & 0.152 & 0.959 & 0.562\\
0.10 & 0.150 & 0.960 & 0.576\\
0.15 & 0.151 & 0.963 & 0.581\\
\end{tabular}
\end{center}
\end{minipage}
}
\vspace{-.4em}
\caption{\textbf{Ablation studies}. All experiments are evaluated on nuScenes validation set. (a) Based on the full 3D supervision, we extra conduct temporal 2D supervision; (b) Based on only 2D supervision of the current frame, we remove the 3D supervision step by step; (c) We study the form of temporal 2D labels, it shows the symmetry supervision and enough temporal interval work better; (d) By randomly jitting the width and height of the projected 2D GT, we study the effect of 3D priors. (Default settings are marked in \colorbox{baselinecolor}{gray}). \label{tab:ablation_study}}
\vspace{-.1in}
\end{table*}

\section{Experiments\label{sec:exp}}

In this section, we conduct our experimental analyses on nuScenes dataset~\cite{nuScenes}. It starts with the ablation studies on dropping 3D supervisions and adding temporal 2D supervisions step by step in Sec.~\ref{sec:exp:drop_add_ablation_study}. Next, we report the performance on hybrid training between 3D and 2D supervision in Sec.~\ref{sec:exp:joint_training}. Finally, we delve into more interesting results and discussions in Sec.~\ref{sec:exp:discussion}.

\subsection{Experimental Setup\label{sec:exp:exp_setup}}

\paragraph{Datasets and Metrics.}
We conduct the exploration study on nuScenes dataset~\cite{nuScenes}. It is a large-scale dataset for automatic driving scenes including 700/150/150 scenes for training, validation, and testing, respectively. Each scene is a continuous 20s video containing 40 keyframes with precise 3D annotations.
It collects multimodal data from LiDAR, camera, and radar, while we only use the camera data. The official provides several evaluation metrics including mean Average Precision (mAP), mean Average Translation Error (mATE), mean Average Scale Error (mASE), mean Average Orientation Error(mAOE), mean Average Velocity Error(mAVE), mean Average Attribute Error(mAAE), and nuScenes Detection Score (NDS). For the full 2D supervision setting, we only report mAP, mATE, and mASE. For the joint training setting, we report all evaluation metrics.

\paragraph{Implementation Details.}
If not specified, we adopt ResNet50~\cite{resnet} initialized from ImageNet~\cite{imagenet} as the backbone for all experiments. Other networks like FPN~\cite{fpn} and head are randomly initialized and trained from scratch in an end-to-end manner. We train the model with SGD optimizer with batch size 16 on 8 GTX 2080Ti GPUs. We set the learning rate to 0.002 and use gradient clip and warm-up strategy for training. The warm-up iterations are set to 500 with the 0.33 ratio. We train the model with 12 epochs for ablation study, while we report the main results with extra 12 epochs of fine-tuning. During training, we use the 2D box projected from the 3D ground truth as our temporal supervision.

\begin{figure*}[t]
\centering
\tablestyle{5pt}{1.2}
\begin{tabular}{cc}
% \hspace{-0.7em}
\includegraphics[width=0.485\textwidth]{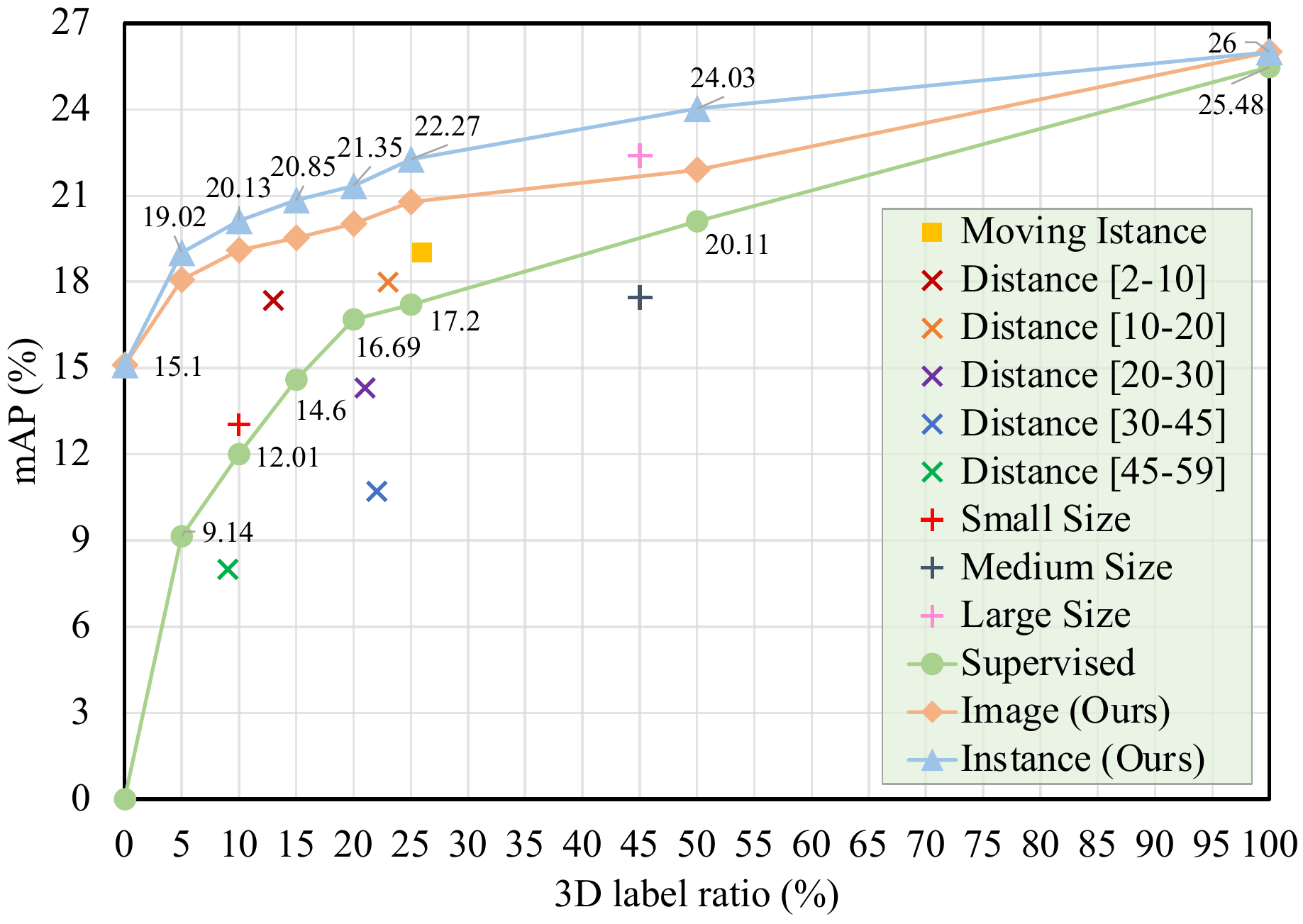} &
\includegraphics[width=0.486\textwidth]{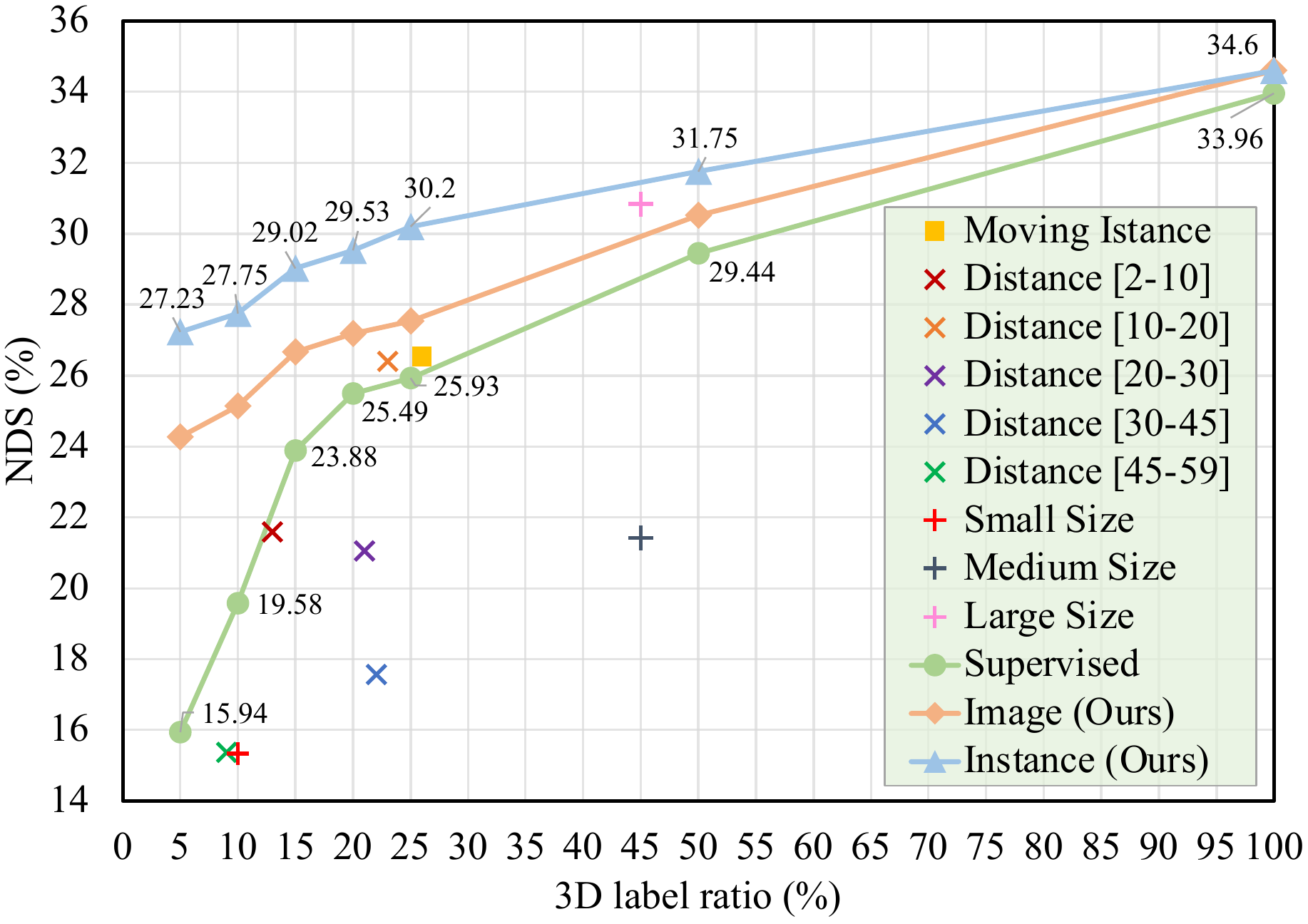}\\
(a) mAP performances with supervision of projected 2D labels & (b) NDS performances with supervision of projected 2D labels \\
\end{tabular}
\vspace{-.4em}
\caption{\textbf{Hybrid training with different ratios between 3D and 2D labels.} Supervised means only using partial 3D labels for training. Image and instance indicate using randomly sampling images and objects for annotating 3D labels. Others use motion state, distance from D1 to D2 meters, and 2D size priors for annotating instances with 3D labels. The size measurement is divided by area the same as COCO~\cite{coco}.}
\label{fig:performance_hybrid_training}
\vspace{-.05in}
\end{figure*}

\begin{table*}[t]
\centering
\tablestyle{7pt}{1.2}
\begin{tabular}{l|ccc|ccccccc}
Method & 3D Ratio & Backbone & Epoch & mAP$\uparrow$ & mATE$\downarrow$ & mASE$\downarrow$ & mAOE$\downarrow$ & mAVE$\downarrow$ & mAAE$\downarrow$ & NDS$\uparrow$ \\
\hline
FCOS3D~\cite{fcos3d} & 100\% & ResNet101-DCN & 12 & 0.295 & 0.806 & 0.268 & 0.511 & 1.315 & 0.170 & 0.372\\ 
FCOS3D~\cite{fcos3d} & 100\% & ResNet101-DCN & Finetune & 0.316 & 0.755 & 0.263 & 0.458 & 1.307 & 0.169 & 0.393\\
\hline
Hybrid (Ours) & 10\% & ResNet101-DCN & 12 & 0.222 & 0.861 & 0.281 & 0.632 & 1.310 & 0.186 & 0.304\\ 
Hybrid (Ours) & 10\% & ResNet101-DCN & Finetune & 0.236 & 0.842 & 0.280 & 0.715 & 1.233 & 0.186 & 0.315\\ 
Hybrid (Ours) & 25\% & ResNet101-DCN & 12 & 0.251 & 0.826 & 0.275 & 0.690 & 1.283 & 0.170 & 0.333\\ 
Hybrid (Ours) & 25\% & ResNet101-DCN & Finetune & 0.266 & 0.813 & 0.270 & 0.643 & 1.280 & 0.179 & 0.345 \\ 
\end{tabular}
\vspace{-.4em}
\caption{\textbf{Main results on hybrid training.} We also report the finetuning performance from the same model with 12 epochs.}
\label{tab:main_result}
\vspace{-.1in}
\end{table*}

\subsection{Ablation Study\label{sec:exp:drop_add_ablation_study}}

We first conduct an incremental experiment for adding the 2D temporal supervision on 3D supervised baseline (oracle). Afterwards, we carry out ablation studies on dropping 3D supervisions and adding temporal 2D supervisions step by step. Finally, a simulation experiment is designed to remove the ambiguity of 2D labels achieved from human annotation or projected from 3D annotations.

\paragraph{Ablation 1: strong oracle.} We start out reporting the full 3D supervised baseline for comparison in Tab.~\ref{tab:strong_oracle}. It achieves 25.5\% mAP and 33.9\% NDS performance. Based on the baseline, we add the 2D supervision for the current frame, which boosts the performance by slightly 0.4\% mAP and 0.5\% NDS. The gains are the same as PGD~\cite{pgd} but we only leverage 2D labels to supervise the projected 2D predictions instead of adding an extra 2D prediction head. By projecting with GT depth for predictions, the performance keeps comparable. When temporal 2D labels participate in supervision, the performance drops slightly, which may be attributed to the biased supervision. Large temporal interval can offset the bias, as maintained in Sec.~\ref{sec:joint_training}.

\paragraph{Ablation 2: removing 3D supervision.} To train the model with only 2D labels, we start out using the 2D center for label assignment mentioned in Sec.~\ref{sec:joint_training} instead of the 3D aware center. Tab.~\ref{tab:remove_3d_supervision} shows slight performance degradation. Next, we drop 3D size and rotation supervision, which largely affects the performance of mASE and mAOE. Then, we further remove the supervision of offset, it barely affects all performance metrics. Specially, the absence of the above 3D supervision cues only bring about a slightly negative impact on mAP and mATE metrics. The phenomenon largely depends on the definition of mAP (\ie, it measures the center distance to calculate average precision), while mATE metric is also measured by center distance error. Finally, we remove the depth supervision, which significantly damages all performance metrics (\ie, 2.6\% mAP, 1.149 mATE, and 0.856 mASE). The result makes sense since 2D labels of single frame will suffer from serious depth information deficiency. It poses a challenge for perceiving the 3D clues.

\paragraph{Ablation 3: effectiveness of the temporal 2D supervision.} After hitting rock bottom in performance, we start studying the effectiveness of adding temporal 2D supervision to estimate the 3D information again. The results reported in Tab.~\ref{tab:temporal_2d_supervision} can be combined into multiple groups for exploring several interesting findings. By comparing A v.s (B, C, D), temporal 2D supervision significantly improves nearly 10\% mAP, while mATE and mASE achieve over 15\% and 30\% gains, respectively. It reflects that using temporal 2D labels by our method is effective to perceive depth information. By comparing (B, C, D) v.s. (E, F, G), it reveals that symmetrical supervisions outperform unilateral supervision, which agrees with our analysis in Sec.~\ref{sec:joint_training}. We further compare performance inside (E, F, G, H), which reflects that the appropriately large temporal interval can constantly boost performance. Once the temporal interval is set too big, the performance begins to drop. The phenomenon also agrees with our analysis in Sec.~\ref{sec:joint_training}. As for other temporal 2D supervision manners, they are the sub-optimal trade-off schemes between optimal \emph{temporal interval} and \emph{symmetrical supervision}. Therefore, we use 2D temporal supervision with [-3, 0, 3] intervals as default setting in subsequent exploration.

\paragraph{Ablation 4: is the 3D prior affect the performance?} We further study the 2D labels projected from 3D ground truth whether triggering 3D information leakage for training. To do this, we employ a recent advanced 2D detector~\cite{yolox} pretrained on COCO~\cite{coco} to predict the 2D boxes of the overall train set. Then, we simply utilize the greedy Intersection-over-Union (IoU) searching strategy to perform the matching between predicted 2D boxes and projected 2D ground truths. It reveals that the size and location shifts are:

\begin{table}[h!]
\centering
\vspace{-.1in}
\tablestyle{13pt}{1.2}
\begin{tabular}{c|ccc}
shifting scale & $\le$ 0.05 & $\le$ 0.10 & $\le$ 0.15\\
\shline
ratio & 18\% & 61\% & 19\% \\
\end{tabular}
\vspace{-.18in}
\end{table}

\begin{table*}[t]
\centering
\tablestyle{7pt}{1.2}
\begin{tabular}{lll|ccccccccccc}
Setting & 3D label & Train & mAP & car & truck & bus & ped. & motor. & bicycle & TC. & barrier & trailer & CV. \\
\hline
Full 3D & 100\% & super. & 0.255 & 0.421 & 0.172 & 0.217 & 0.373 & 0.224 & 0.212 & 0.467 & 0.375 & 0.047 & 0.041\\ 
Random  & 16\% & super. & 0.152 & 0.295 & 0.056 & 0.051 & 0.280 & 0.076 & 0.119 & 0.349 & 0.243 & 0.002 & 0.008\\
Full 2D & 0\% & weak & 0.151 & 0.218 & 0.067 & 0.017 & 0.109 & 0.150 & 0.133 & 0.397 & 0.350 & 0.004 & 0.028\\[.5mm]
\end{tabular}
\vspace{-.6em}
\caption{\textbf{Different categories.} ``Random'' indicates randomly sampling 16\% images with 3D supervision.
\label{tab:perfomance_different_categories}}
\vspace{-.15in}
\end{table*}

\begin{table}[t]
\centering
\tablestyle{4pt}{1.2}
\begin{tabular}{ll|ccccc}
Setting & 3D label & 2$\sim$10 & 10$\sim$20 & 20$\sim$30 & 30$\sim$45 & 45$\sim$59\\
\hline
Full 3D sup. & 100\% & 0.449 & 0.363 & 0.217 & 0.078 & 0.023\\ 
Random sup. & 16\%& 0.305 & 0.218 & 0.107 & 0.029 & 0.009\\
Full 2D weak & 0\% & 0.356 & 0.227 & 0.098 & 0.019 & 0.003 \\[.5mm]
\end{tabular}
\vspace{-.6em}
\caption{\textbf{Different distances}. The distance is measured in meters.
\label{tab:perfomance_different_distance}}
\vspace{-.15in}
\end{table}

\noindent We note that the shifting scale of 98\% of predicted 2D boxes is smaller than 0.15. Once given the ground truth 2D annotations, the shifting scale will be smaller than 0.15 too. Based on the shifting estimation, we study the effect of 3D prior. Tab.~\ref{tab:remove_3d_prior} reports the performance by randomly jitting the scale of projected 2D GTs. It shows that removing the 3D prior rarely affects the performance, revealing the projected 2D labels can not trigger the  3D information leakage.

\subsection{Hybrid Training\label{sec:exp:joint_training}}

Given partial 2D supervision and 3D labels, we study the performance of our hybrid training framework. We conduct extensive experiments on the training set of nuScenes dataset and ablate the ratio between 3D and 2D labels. 

\paragraph{Random split on images and instances.}
We randomly split the dataset based on images and instances before experiments. 
As shown in Fig.~\ref{fig:performance_hybrid_training}, it shows that our hybrid training can constantly improve the mAP and NDS performance by replacing the 3D supervision with temporal 2D ones. Only given 5\% 3D labels, our method can boost the mAP and NDS performance nearly twice. While providing 25\% 3D labels, our method achieves 87.5\% mAP and 88.9\% NDS performances of the full 3D supervision. In addition, it reveals that the random instance split manner performs better than the image split one. It introduces a practical insight to only annotate partial 3D GTs. Workers can select easier instances to annotate, enabling them to reduce massive costs.

\paragraph{Could the split priors on motion state, distance, and size work better?}
We further study employing priors on the instance division. Fig.~\ref{fig:performance_hybrid_training} reveals that only providing 3D labels for moving instances can achieve improvement but underperform the random split manner. As for the distance and size split manners, only the large size prior shows better performance compared with the random image manner. Unexpectedly, other split priors like remote distance and medium size bring about performance degradation. These interesting findings reveal that random annotation works well, which may introduce a better distribution between the 3D and 2D supervision for optimization.

\paragraph{Main Results.}
Tab.~\ref{tab:main_result} report the performances on large backbone ResNet101 with DCN~\cite{dcn}. It demonstrates that our hybrid training framework can improve all evaluation metrics. When training with both 12 epochs and continually fine-tuning (the same as FCOS3D~\cite{fcos3d}), our method can achieve promising performance with small ratio of 3D labels.

\subsection{More Interesting Results and Discussions\label{sec:exp:discussion}}

\paragraph{Performance on different categories and distances.}
Tab~\ref{tab:perfomance_different_categories} shows traffic cone and barrier classes achieve unexpected performances (\ie, only 7.0\% and 2.5\% mAP smaller than full 3D supervision) with full 2D supervision.
Since these two classes are stationary objects, the temporal 2D supervision will not trigger the bias loss. 
However, the pedestrian and bus classes perform badly compared with full 3D supervision.
We deem it attributes to \emph{larger or smaller 3D sizes}.
In the motion state, the minimum size class, \ie, pedestrians, will trigger a large location shift deviated from the homography warping transformation.
As for the large size class like buses and trucks, they are prone to encounter mutation of size due to the drastic visual surface changing. Through the fair comparison between the random sample (within image) 3D supervision and the fully 2D one, we find that the former predicts the moving objects well and the latter perceives stationary objects better.
See Tab.~\ref{tab:perfomance_different_distance}, we also report the performance on different distances. It indicates that temporal 2D supervision predicts nearby objects well, which is lower than full 3D supervision by 9.4\% mAP. While the distant objects are poorly estimated mainly due to the lack of depth.

\paragraph{Discussion and limitation.}
Our method shows promising potential for training 3D visual detectors with large-scale cheap 2D labels. The overall pipeline is simple and does not need the point cloud data. In addition, the full 2D supervision manner will liberate the annotation procedure since it does not require any point clouds for reference.

One nonnegligible limitation of our work is the motion ambiguity under the homography warping procedure. We introduce symmetry supervision to alleviate the issue, but it is far from enough. It poses a significantly valuable but challenging topic for future work.

% \vspace{.02in}
\section{Conclusion\label{sec:conclusion}}
In this paper, we present the visual 3D detector supervised by massive 2D labels and a few 3D ones. With the temporal 2D supervision in our training pipeline, any 3D detectors can effectively estimate the 3D information with 2D annotations. The comprehensive experiments show promising results. We hope our exploration can provide new insights for using a large number of 2D annotations for 3D perception.

{\small
\bibliographystyle{ieee_fullname}
\bibliography{cite}

\begin{thebibliography}{10}\itemsep=-1pt

\bibitem{megnet}
Gwangbin Bae, Ignas Budvytis, and Roberto Cipolla.
\newblock Multi-view depth estimation by fusing single-view depth probability
  with multi-view geometry.
\newblock In {\em CVPR}, 2022.

\bibitem{nuScenes}
Holger Caesar, Varun Bankiti, Alex~H. Lang, Sourabh Vora, Venice~Erin Liong,
  Qiang Xu, Anush Krishnan, Yu Pan, Giancarlo Baldan, and Oscar Beijbom.
\newblock nuscenes: {A} multimodal dataset for autonomous driving.
\newblock In {\em CVPR}, 2020.

\bibitem{renderer}
Wenzheng Chen, Huan Ling, Jun Gao, Edward Smith, Jaakko Lehtinen, Alec
  Jacobson, and Sanja Fidler.
\newblock Learning to predict 3d objects with an interpolation-based
  differentiable renderer.
\newblock In {\em NeurIPS}, 2019.

\bibitem{adascale}
Ting-Wu Chin, Ruizhou Ding, and Diana Marculescu.
\newblock Adascale: Towards real-time video object detection using adaptive
  scaling.
\newblock {\em Proceedings of Machine Learning and Systems}, 2019.

\bibitem{dcn}
Jifeng Dai, Haozhi Qi, Yuwen Xiong, Yi Li, Guodong Zhang, Han Hu, and Yichen
  Wei.
\newblock Deformable convolutional networks.
\newblock In {\em ICCV}, 2017.

\bibitem{fischler1981random}
Martin~A Fischler and Robert~C Bolles.
\newblock Random sample consensus: a paradigm for model fitting with
  applications to image analysis and automated cartography.
\newblock {\em Communications of the ACM}, pages 381--395, 1981.

\bibitem{ota}
Zheng Ge, Songtao Liu, Zeming Li, Osamu Yoshie, and Jian Sun.
\newblock Ota: Optimal transport assignment for object detection.
\newblock In {\em CVPR}, 2021.

\bibitem{yolox}
Zheng Ge, Songtao Liu, Feng Wang, Zeming Li, and Jian Sun.
\newblock Yolox: Exceeding yolo series in 2021.
\newblock {\em arXiv preprint arXiv:2107.08430}, 2021.

\bibitem{sul}
Georgia Gkioxari, Nikhila Ravi, and Justin Johnson.
\newblock Learning 3d object shape and layout without 3d supervision.
\newblock In {\em CVPR}, 2022.

\bibitem{resnet}
Kaiming He, Xiangyu Zhang, Shaoqing Ren, and Jian Sun.
\newblock Deep residual learning for image recognition.
\newblock In {\em CVPR}, 2016.

\bibitem{kato2018neural}
Hiroharu Kato, Yoshitaka Ushiku, and Tatsuya Harada.
\newblock Neural 3d mesh renderer.
\newblock In {\em CVPR}, 2018.

\bibitem{imagenet}
Alex Krizhevsky, Ilya Sutskever, and Geoffrey~E Hinton.
\newblock Imagenet classification with deep convolutional neural networks.
\newblock {\em Communications of the ACM}, 2017.

\bibitem{pointpillars}
Alex~H Lang, Sourabh Vora, Holger Caesar, Lubing Zhou, Jiong Yang, and Oscar
  Beijbom.
\newblock Pointpillars: Fast encoders for object detection from point clouds.
\newblock In {\em CVPR}, 2019.

\bibitem{bevstereo}
Yinhao Li, Han Bao, Zheng Ge, Jinrong Yang, Jianjian Sun, and Zeming Li.
\newblock Bevstereo: Enhancing depth estimation in multi-view 3d object
  detection with dynamic temporal stereo.
\newblock {\em arXiv preprint arXiv:2209.10248}, 2022.

\bibitem{fpn}
Tsung-Yi Lin, Piotr Doll{\'a}r, Ross Girshick, Kaiming He, Bharath Hariharan,
  and Serge Belongie.
\newblock Feature pyramid networks for object detection.
\newblock In {\em CVPR}, 2017.

\bibitem{retinanet}
Tsung-Yi Lin, Priya Goyal, Ross Girshick, Kaiming He, and Piotr Doll{\'a}r.
\newblock Focal loss for dense object detection.
\newblock In {\em ICCV}, 2017.

\bibitem{coco}
Tsung-Yi Lin, Michael Maire, Serge Belongie, James Hays, Pietro Perona, Deva
  Ramanan, Piotr Doll{\'a}r, and C~Lawrence Zitnick.
\newblock Microsoft coco: Common objects in context.
\newblock In {\em CVPR}, 2014.

\bibitem{weak3}
Haizhuang Liu, Huimin Ma, Yilin Wang, Bochao Zou, Tianyu Hu, Rongquan Wang, and
  Jianshen Chen.
\newblock Eliminating spatial ambiguity for weakly supervised 3d object
  detection without spatial labels.
\newblock In {\em ACM MM}, 2022.

\bibitem{liu2019soft}
Shichen Liu, Weikai Chen, Tianye Li, and Hao Li.
\newblock Soft rasterizer: Differentiable rendering for unsupervised
  single-view mesh reconstruction.
\newblock {\em arXiv preprint arXiv:1901.05567}, 2019.

\bibitem{weak1}
Qinghao Meng, Wenguan Wang, Tianfei Zhou, Jianbing Shen, Luc~Van Gool, and
  Dengxin Dai.
\newblock Weakly supervised 3d object detection from lidar point cloud.
\newblock In {\em ECCV}, 2020.

\bibitem{WeakM3D}
Liang Peng, Senbo Yan, Boxi Wu, Zheng Yang, Xiaofei He, and Deng Cai.
\newblock Weakm3d: Towards weakly supervised monocular 3d object detection.
\newblock In {\em ICLR}, 2022.

\bibitem{giou}
Hamid Rezatofighi, Nathan Tsoi, JunYoung Gwak, Amir Sadeghian, Ian Reid, and
  Silvio Savarese.
\newblock Generalized intersection over union: A metric and a loss for bounding
  box regression.
\newblock In {\em CVPR}, 2019.

\bibitem{scharstein2002taxonomy}
Daniel Scharstein and Richard Szeliski.
\newblock A taxonomy and evaluation of dense two-frame stereo correspondence
  algorithms.
\newblock {\em IJCV}, 2002.

\bibitem{pointrcnn}
Shaoshuai Shi, Xiaogang Wang, and Hongsheng Li.
\newblock Pointrcnn: 3d object proposal generation and detection from point
  cloud.
\newblock In {\em CVPR}, 2019.

\bibitem{fcos}
Zhi Tian, Chunhua Shen, Hao Chen, and Tong He.
\newblock Fcos: Fully convolutional one-stage object detection.
\newblock In {\em ICCV}, 2019.

\bibitem{pgd}
Tai Wang, ZHU Xinge, Jiangmiao Pang, and Dahua Lin.
\newblock Probabilistic and geometric depth: Detecting objects in perspective.
\newblock In {\em CoRL}, 2022.

\bibitem{fcos3d}
Tai Wang, Xinge Zhu, Jiangmiao Pang, and Dahua Lin.
\newblock Fcos3d: Fully convolutional one-stage monocular 3d object detection.
\newblock In {\em ICCV}, 2021.

\bibitem{manydepth}
Jamie Watson, Oisin Mac~Aodha, Victor Prisacariu, Gabriel Brostow, and Michael
  Firman.
\newblock The temporal opportunist: Self-supervised multi-frame monocular
  depth.
\newblock In {\em CVPR}, 2021.

\bibitem{FGR}
Yi Wei, Shang Su, Jiwen Lu, and Jie Zhou.
\newblock {FGR:} frustum-aware geometric reasoning for weakly supervised 3d
  vehicle detection.
\newblock In {\em ICRA}, 2021.

\bibitem{second}
Yan Yan, Yuxing Mao, and Bo Li.
\newblock Second: Sparsely embedded convolutional detection.
\newblock {\em Sensors}, 18(10):3337, 2018.

\bibitem{stream}
Jinrong Yang, Songtao Liu, Zeming Li, Xiaoping Li, and Jian Sun.
\newblock Real-time object detection for streaming perception.
\newblock In {\em CVPR}, 2022.

\bibitem{dbqssd}
Jinrong Yang, Lin Song, Songtao Liu, Zeming Li, Xiaoping Li, Hongbin Sun, Jian
  Sun, and Nanning Zheng.
\newblock Dbq-ssd: Dynamic ball query for efficient 3d object detection.
\newblock {\em arXiv preprint arXiv:2207.10909}, 2022.

\bibitem{3dssd}
Zetong Yang, Yanan Sun, Shu Liu, and Jiaya Jia.
\newblock 3dssd: Point-based 3d single stage object detector.
\newblock In {\em CVPR}, 2020.

\bibitem{mvsnet}
Yao Yao, Zixin Luo, Shiwei Li, Tian Fang, and Long Quan.
\newblock Mvsnet: Depth inference for unstructured multi-view stereo.
\newblock In {\em ECCV}, 2018.

\bibitem{centerpoint}
Tianwei Yin, Xingyi Zhou, and Philipp Krahenbuhl.
\newblock Center-based 3d object detection and tracking.
\newblock In {\em CVPR}, 2021.

\bibitem{atss}
Shifeng Zhang, Cheng Chi, Yongqiang Yao, Zhen Lei, and Stan~Z Li.
\newblock Bridging the gap between anchor-based and anchor-free detection via
  adaptive training sample selection.
\newblock In {\em CVPR}, 2020.

\bibitem{aissd}
Yifan Zhang, Qingyong Hu, Guoquan Xu, Yanxin Ma, Jianwei Wan, and Yulan Guo.
\newblock Not all points are equal: Learning highly efficient point-based
  detectors for 3d lidar point clouds.
\newblock In {\em CVPR}, 2022.

\bibitem{centernet}
Xingyi Zhou, Dequan Wang, and Philipp Kr{\"a}henb{\"u}hl.
\newblock Objects as points.
\newblock {\em arXiv preprint arXiv:1904.07850}, 2019.

\bibitem{voxelnet}
Yin Zhou and Oncel Tuzel.
\newblock Voxelnet: End-to-end learning for point cloud based 3d object
  detection.
\newblock In {\em CVPR}, 2018.

\end{thebibliography}
}

\end{document}